# Augmenting Gastrointestinal Health: A Deep Learning Approach to Human Stool Recognition and Characterization in Macroscopic Images

**David Hachuel**
Cornell Tech
Cornell University
New York, NY, USA
dh649@cornell.edu

**Akshay Jha**
Cornell Tech
Cornell University
New York, NY, USA
aj545@cornell.edu

**Deborah Estrin, PhD**
Cornell Tech
Cornell University
New York, NY, USA
destrin@cs.cornell.edu

**Alfonso Martinez**
Massachusetts Institute of Technology
Cambridge, MA, USA
almartin@mit.edu

**Kyle Staller, MD, MPH**
Massachusetts General Hospital
Boston, MA, USA
kstaller@mgh.harvard.edu

**Christopher Velez, MD**
Massachusetts General Hospital
Boston, MA, USA
cvelez@partners.org

**ABSTRACT**

UPDATED—4 December 2018. **Purpose** - Functional bowel diseases, including irritable bowel syndrome, chronic constipation, and chronic diarrhea, are some of the most common diseases seen in clinical practice. Many patients describe a range of triggers for altered bowel consistency and symptoms. However, characterization of the relationship between symptom triggers using bowel diaries is hampered by poor compliance and lack of objective stool consistency measurements. We sought to develop a stool detection and tracking system using computer vision and deep convolutional neural networks (CNN) that could be used by patients, providers, and researchers in the assessment of chronic gastrointestinal (GI) disease. **Methods** - We collected 938 human stool images from anonymous sources, which were then examined by 2 expert GI motility physicians who rated the corresponding consistency level using the Bristol Stool Scale (BSS), a known correlate of colonic transit time. Agreement between raters was calculated using Cohen's Kappa statistic. In addition, we generated a secondary dataset of 36000 images of synthetic stool samples made of play-doh for comparison purposes. First, we investigated specimen detection using convolutional autoencoders, a variant of CNN architectures, and, in particular, a modification of SegNet, a known image segmentation architecture. We used mean intersection-over-union (mIoU), also known as Jaccard index, to evaluate the similarity between the area recognized by the SegNet and the actual specimen location. We then trained a deep residual CNN (ResNet) to classify each specimen's consistency based on the BSS. We defined the classification accuracy as the ratio of correct predictions to the total number of predictions made by the ResNet. **Results** - In terms of inter-rater reliability, the Kappa score was $\kappa = 0.5840$. Regarding the stool detection task, our modified SegNet produced a 71.93% mIoU on a test set of 282 images when trained on 651 images. Current state-of-the-art segmentation models reach 83.2% mIoU when trained on significantly larger datasets. As for BSS classification using the ResNet architecture, our mean accuracy is 74.26% on a test set of 272 images when trained on 614 images. In comparison, the ResNet trained on the secondary dataset achieved 99.4% accuracy in classifying BSS on a test set of 10800 images when trained on 25200 images. **Conclusion** - This pilot study demonstrates that deep neural networks can provide accurate detection and descriptive characterization of stool images. With a larger stool image dataset, we expect this method's performance to improve as seen when trained on our secondary dataset. This method could be applicable across a variety of clinical and research settings to help patients and providers collect accurate information on bowel movement habits and improve disease diagnosis and management.

**ACM Classification Keywords**

H.5.m. Information interfaces and presentation (e.g., HCI): Miscellaneous; See http://acm.org/about/class/1998 for the full list of ACM classifiers. This section is required.

**INTRODUCTION**

Chronic digestive disorders such as irritable bowel syndrome (IBS) affect a significant portion of the US population. About 20% of Americans suffer from IBS [3], one of the most common of these conditions and considered an important public health problem. Each IBS patient is unique in its symptoms and in its potential sensitivity to life inputs such as diet, menstruation, mental and physical activity. For this reason, providers currently try to identify flare-up triggers by analyzing paper or digital diaries and look for correlations between food, activity and symptoms [2]. However, patient journals are often paper-based, incomplete, mixed up and/or inaccurate. This makes the providers job quite difficult in addition to the short consultation time they have to make sense of the collected data. It is no surprise then that a study by O'Sullivan et al found that about 77% of IBS patients and 56% of Inflammatory Bowel Disease (IBD) were dissatisfied with the feedback they received from providers and required further information about their condition [8].

During the past decade, a decrease in the cost of computational resources and sensing hardware has given rise to an explosion of the quantified-self movement. This movement, also known as lifelogging, began in the early 2000's and tries to incorporate new technology and sensors that capture as many aspects of our daily lives as possible. The movement was coined by Gary Wolf and Kevin Kelly of Wired magazine who foretold that, with ubiquitous sensing, everyday changes could become the object of detailed analysis and thus creating a new kind of knowledge [14]. As digital health applications create new dynamics between patients and providers, emerging health monitoring technologies change the patient-provider dynamics. We propose a novel intervention to improve research and care delivery in the gastroenterology setting.

**RELATED WORK**

In a 2016 study [10], researchers evaluated the feasibility and usability of food and GI journal smartphone apps in IBS management as an alternative to paper diaries given their unreliability. The authors recruited 11 participants and asked them to track food and GI symptoms over the course of 2 weeks using a mobile app.

In terms of feasibility, researchers found that on average participants had a daily completion rate of 112% (overreporting) and 78% for meal and symptoms respectively. This suggests that participants found it more difficult to log symptoms than to log meal consumption. In addition, participants displayed a decrease in daily completion rates of 25% and 17% for meal and symptoms respectively from week 1 to week 2 of the study. Regarding usability, the researchers used the System Usability Scale (SUS), a 10-item questionnaire with 5 response options that is commonly used to measure usability and that is technology independent. On a score range from 65 to 97.5 and mean 68, the researchers found their app scored above average at 83. When the researchers added notifications to the app, the SUS rose to 92. However, when measuring improvements in symptoms severity, the researchers found no statistically significant decreases. On a qualitative note, the participants noted that they wished the app could provide "answers" or at least better data visualizations with analysis. They seem to want guidance on how to change their habits. Overall, participants found the app to be useful but there seem to be limitations in the researcher's methodology and approach. First, a sample size of 11 is not representative enough of 60 million Americans with IBS and may be conditioned by localities. Second, the study only covered a 2-week period which begs the question of what would have happened to the already decreasing compliance rate had the study continued. Although the researchers note the limitations of their study, they point out that electronic journals are in general better than paper-based ones resulting in improved compliance, higher quality data and easier data handling.

Current research seems to suggest that patients with chronic digestive disorders and their providers should have access to more efficient, standardized, actionable and effective ways to collect and share life-log data about their habits and symptoms. Diagnostic self-tracking, tracking to answer a specific question, could enable them to identify specific lifestyle modifications that could reduce symptom severity [2]. In 2017, Karkar et al developed *TummyTrials*, a self-experimentation app to support patients suffering from IBS [5]. *TummyTrials* uses single-case experimental designs (SCD) to help patients in designing, executing and analyzing self-experiments to detect specific foods that trigger symptoms.

In the evaluation of the effectiveness of *TummyTrials*, the researchers recruited a group of 15 participants who had overall positive experiences with lower burden and higher compliance through self-accountability reporting that the app provided structure, discipline and reminders. In terms of usability and user burden, participants reported an above-average mean SUS of 83 and a mean User Burden Scale (UBS) grade between B and C. In comparison to other apps, *TummyTrials* participants reported higher compliance rates (100% compliance for 12 out of 15 participants).

Self-tracking for chronic GI patients is a difficult and burdensome task. What is more, clinicians don't receive the necessary training to interpret the non-standardized journal data nor they have enough time to do so during consultations. Some interventions like *TummyTrials* try to motivate and empower the patients by simplifying data collection process and providing a clear structure for self-experimentation. However, there are existing limitations around the precision and objectivity of patients' self-reports. For instance, stool consistency is a critical component in the description of normal or altered bowel habits and the Bristol Stool Form Scale (BSFS) is a clinically validated scale used extensively in clinical practice [1]. Yet, studies suggest patient-reported accounts of stool consistency do not always agree with clinical assessment. In a series of experiments, Blake et al collected 59 real stools provided by volunteers and asked experts in GI research to classify the BSFS. The researchers then asked the volunteers to provide the BSFS classification as well. In comparison, only 36% of stools were correctly assigned to the same BSFS type classified by the experts, and 75% were classified with at most 1 BSFS level deviation from the expert rating. Stool consistency is a major component of GI symptoms tracking but it is currently being captured in a subjective way which might lead to wrong conclusions.

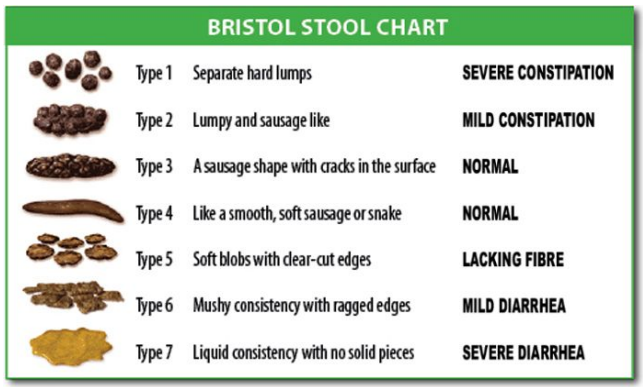

Figure 1. Bristol Stool Form Scale Illustration.

In another study published in Neurogastroenterology & Motility [4], a group of researchers found great differences in patients' perceptions of diarrhea and constipation. After asking 70 IBS patients (Rome III criteria) to rate dissatisfaction with stool consistency, they found no statistically significant correlation (Pearson correlation of 0.26 for IBS-Diarrhea and 0.47 for IBS-Constipation) between the self-reported dissatisfaction and an objective Fecal Water Content (FWC) analysis, an indicator of consistency.

## RESEARCH QUESTION

Nowadays, it is hard to ignore the omnipresence of intelligent algorithms in our everyday experience. From step trackers and recommendation systems, to voice assistants and self-driving cars, these statistical or machine learning methods augment our human capabilities to an unprecedented level. These methods are also becoming good at sight or the ability to understand images. Take for instance the newest iPhone models and their innovative face detection system. Similar image processing approaches can be used to characterize medical images such as x-rays or magnetic resonance imaging scans. Why not use this approach to analyze images of stools and thus bring the expert "eyes" to patients' homes?

We expect that such approach would remove the subjectivity in current patient reports, make a more precise measurement and minimize the user experience burden. Such an intervention could also perfect the diagnosis of functional GI disorders (FGIDs) such as IBS which currently relies of subjective symptom reports, thus transforming macroscopic stool characteristics into a biomarker. In this paper, we developed a 2-part system to recognize human stool specimens and characterize the BSFS in macroscopic images.

## IMAGE DATASET

We now describe how the data was collected as well as annotated by expert humans.

### Image Collection

To our knowledge and at the time of writing, there are no stool image datasets available to use. Our data collection strategy was 2-fold. First, we scraped the web for images that were shared in platforms such as *Reddit*. Second, we created a simple smartphone application called *Train augGI* (http://train.auggi.ai) that allows anyone to anonymously upload an image (see Figure 2). In total, we collected 938 images.

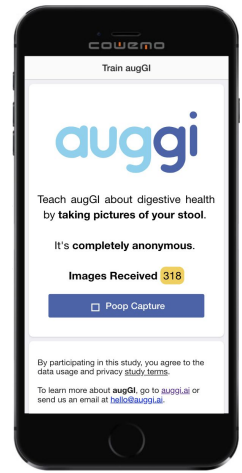

Figure 2. *Train augGI* Smartphone App Screenshot.

### Image Annotation

We collaborated with 2 expert GI motility physicians who rated all images based on the 7-level BSFS. This image annotation process was performed using an online collaborative platform called *LabelBox* (www.labelbox.com). Given the small number of samples, we further consolidated these ratings into 3 groups based on the input from our physician collaborators: "constipation" (BSFS 1, 2), "normal" (BSFS 3, 4, 5) and "loose" (BSFS 6, 7). In terms of inter-rater reliability, we calculated Cohen's Kappa statistic producing a moderate score of $\kappa = 0.5840$.

We further performed pixel-wise annotation of the specimen in the images required to train the stool detection model. Pixel-wise annotations produce detailed masks pointing out the exact location of the specimen (see Figure 3).

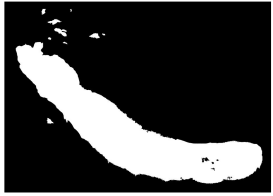

**Figure 3. Specimen Mask Resulting From Pixel-wise Annotation.**

### Dataset Quality Overview

Given the nature of the data collection process, we are aware of certain limitations around data quality. For instance, in a sample set of 100 images, we found 7% were images of stools without a toilet, 29% displayed occlusions caused by reflections, 22% contained toilet paper causing occlusions and 7% contained other irrelevant items in the image.

Moreover, all images were taken with different camera sensors, under different lighting conditions and at different positions or distances. For instance, the average ratio of specimen area to the total number of pixels in an image is 15%. In terms of pixel-wise annotation, it was quite difficult to account for all fecal matter in certain images leading to a non-perfect ground-truth annotation.

Finally, the BSFS ratings are not uniformly distributed. Even if its distribution is representative of the population, when it comes to training a classifier, this is problematic due to class imbalance (see Figure 4). In other words, the classifier might be biased to some classes that are more frequent in the training set.

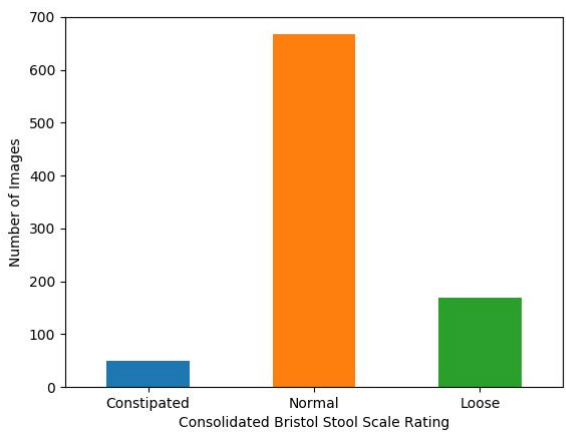

**Figure 4. Distribution of the Consolidated BSFS Ratings.**

### Secondary Dataset

Given the low-quality real image dataset, we decided we needed a benchmark dataset. By injecting some creativity, we resorted to make a secondary stool image dataset using play doh. To do this, we created multiple stool samples in different shapes, positions and amount following levels 1 through 5 of the BSFS scale (see Figure 1).

### Data Augmentation

A common approach to increase variability in the training set and reduce algorithm bias in low-quality datasets is called data augmentation. This involves applying random transformations to the input training images so as to increase entropy. These transformations can be of the following types: skews, rotations, flips, occlusions/erasings, crops/zooms (see Figure 5).

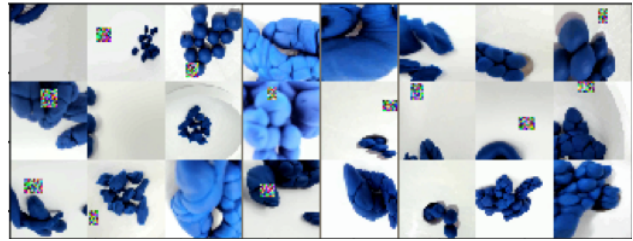

**Figure 5. Random Transformations Applied to the Secondary Dataset.**

## MODEL THEORY AND SELECTION

### Image Classification

The fundamental idea behind image classification is that given an image we can assign a series of labels to that image with certain probability. For humans, recognizing objects in our field of view is a natural ability with which we are born and execute effortlessly. As such, when we see a scene or image, we are able to characterize it and given each object a label without even thinking consciously about it. For a computer, an image is an ordered set or array of pixel values. The latter represent the intensity of each pixel as an integer between 0 and 255 in the gray-scale image or three integers representing red, green and blue (RGB) color channels in the case of a color image (there exist other color-space representations as well). Therefore, the input to an image classification method is an array of dimensions length x height x channels and the output an array of possible labels together with a probability.

Now that we know the inputs and outputs of such an algorithm, we want the latter to find a mapping between the image space and the label space. Images are high-dimensional representations of objects which contain information arranged in 2-dimensional space. Unlike traditional machine learning classifiers such as logistic regression, random forests or even regular neural networks, convolutional neural networks account for the spatial arrangement of local features. This is of course crucial to derive meaningful features and is the reason why these types of network architectures are a good fit for classification and detection in images.

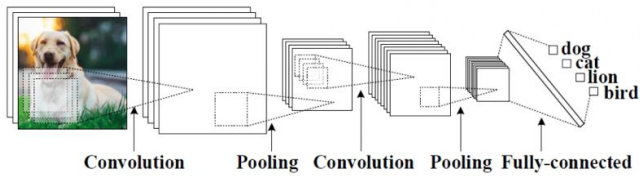

**Figure 6. Illustration of a CNN Architecture Used for Image Classification.**

In general, state-of-the-art neural networks have a large number of parameters that require optimization. In the case of convolutional neural networks, the deeper the network the better it usually performs. But deep architectures come to a high cost in terms of training time and the wealth of well-annotated data required to perform well. For this reason, very few people or organizations build and/or train their own convolutional neural networks. Instead, they leverage something called transfer learning. This refers to the practice of leveraging large pre-trained convolutional neural networks that have already been trained on large image datasets such as ImageNet [15], which contains about 14 million annotated images. These networks have already been taught to extract features from images, i.e. to see, and it becomes significantly simpler to apply these pre-trained parameters, given some level of parameter fine-tuning, to a different dataset or application. We therefore chose to work with a pre-trained convolutional neural network.

**Object Detection**

The motivation behind image object detection is to devise an algorithm that when presented with an image, it can not only point out the type or class of objects but also locate them (see Figure 7).

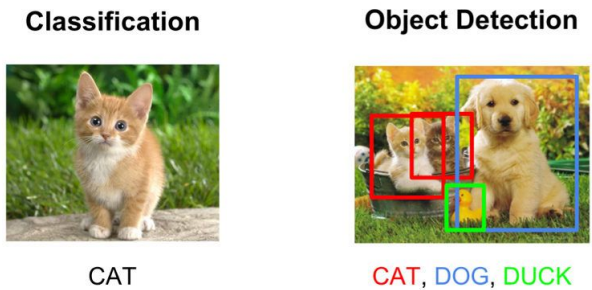

**Figure 7. Difference Between Classification and Object Detection.**

Most detection algorithms are able to draw a bounding box around the detected object instance. Although this is generally useful, these bounding boxes are not pixel-precise contain more than just the desired object, i.e. some surrounding information. Given that our task is to characterize the stool specimen in an uncontrolled environment (i.e. the patient's toilet), the ideal algorithm would precisely locate the specimen in the image and remove all surrounding irrelevant information. In computer vision, image segmentation is the process of partitioning an image into multiple non-overlapping segments or regions of pixels representing different concepts or objects. Since finding precise contours is a key step towards better image characterization [17], we chose segmentation algorithms to perform the detection task with precision.

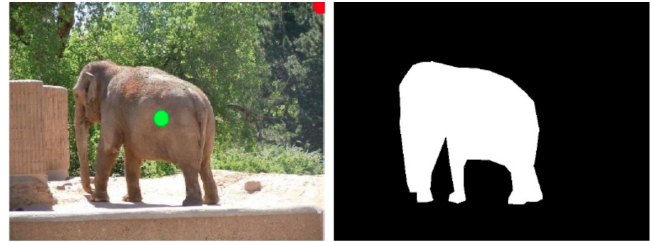

**Figure 8. Example of Image Segmentation Precisely Detecting the Contour of Some Object of Interest.**

**EXPERIMENTAL SETUP**

**Specimen Classification**
For the classification task, we selected one of the simplest, most popular existing configurations called ResNet [18]. The name makes reference to residual learning framework which allows more efficient training for deeper networks given that it prevents vanishing gradient problems by shortcutting connections between layers. *ResNet18* is the shallowest architecture. As its name specifies, it contains 18 layers.

The classification loss was calculated using cross entropy and its accuracy was defined as the ratio of correct predictions to the total number of predictions made by the ResNet. In this case, the classes are "constipation", "normal" and "loose". We trained the *ResNet18* on 614 images which were resize to 224x224 pixels. We trained the model for 30 epochs using stochastic gradient descent as optimizer with momentum and a decaying learning rate of 0.001. We calculated the classification accuracy on a test set of 272 images.

**Specimen Detection**
As for the detection/segmentation task, we chose SegNet, a deep convolutional encoder-decoder architecture that can produce robust image pixel-wise segmentations [15]. All layers in this architecture are convolutional/deconvolutional with batch normalization and rectified linear activations for non-linearity. We introduced a slight modification in the SegNet by adding a sigmoid activation in the last layer consistent with the desired binary output (i.e. stool or not stool). In addition, we loaded pre-trained weights from the VGG-16 pre-trained architecture to take advantage of transfer learning [20].

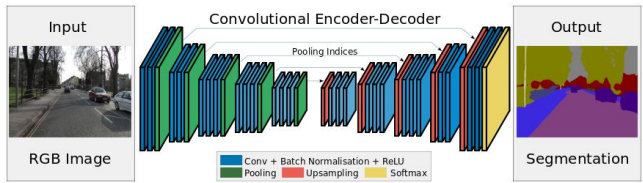

**Figure 9. SegNet Architecture Diagram [15].**

The detection/segmentation loss was calculated using binary cross entropy and its accuracy was evaluated using the mean intersection-over-union (mIoU) representing the similarity between the stool contour/area recognized by the SegNet and the actual specimen location (see Figure 10).

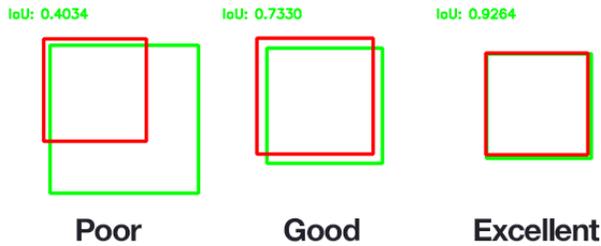

**Figure 10. Illustration of Intersection-Over-Union Measure.**

Training was performed on 651 images which were resized to 224x224 pixels for a duration of 100 epochs and a decaying learning rate of 0.05 with momentum. The training target was a binary mask where the area covered by the specimen was marked as 1 with 0 everywhere else. As recommended by the authors, we trained the SegNet using stochastic gradient descent. We calculated accuracy on a test set of 282 images. Moreover, we experimented by adding 5000 non-stool images from the Coco 2017 validation dataset to test the discriminatory power and robustness of our method [19].

## RESULTS

### Specimen Detection

Overall specimen detection using *SegNet* was strong compared to current state-of-the-art models which reach 83.2% mIoU when trained in datasets like Coco or ImageNet (see Table 1).

| Model Setup | Loss | mIoU |
|---|---|---|
| *SegNet* | 0.1336 | 82.10% |
| *SegNet* with Augmentation | 0.1368 | 71.93% |
| *SegNet* with Coco | 0.2490 | 2.95% |

**Table 1. Comparison of *SegNet* Test Performance in Different Setups.**

Even though mIoU was the highest when training without data augmentation, we noticed an overfitting problem where the train loss was significantly smaller than the test loss (see Figure 11).

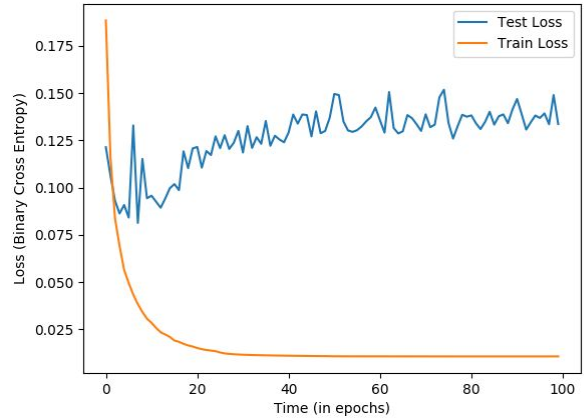

**Figure 11. Loss Trends for *SegNet*.**

When training using data augmentation, we observed no overfitting while still maintaining a relatively high mIoU in test (see Figure 12).

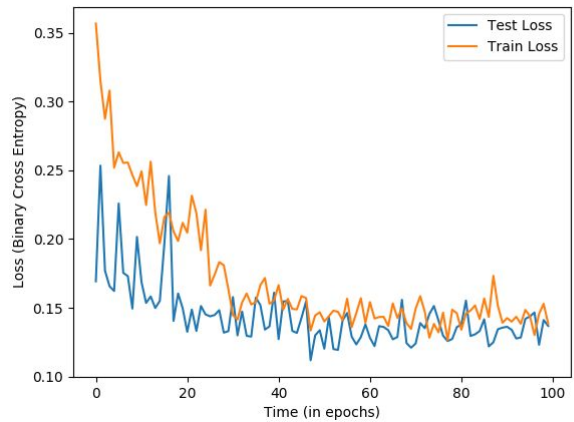

**Figure 12. Loss Trends for *SegNet* using Data Augmentation.**

We also noticed the model had lower mIoU scores for images where the specimen had a small coverage or where other items were causing occlusions. This is consistent with the data quality limitations mentioned previously. For close enough specimens with little occlusion, the *SegNet* performed adequately (see Figure 13).

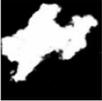

**Figure 13. Sample Segmentation Predictions on Test Images using *SegNet*.**

## Specimen Classification

Due to the limitations in terms of data annotation and class imbalance (low occurrence of certain classes), we observe moderate classification accuracy (see Table 2).

| Model Setup | Accuracy |
| --- | --- |
| *ResNet18* Partial Fine-tuning | 73.53% |
| *ResNet18* Full Fine-tuning | 73.53% |
| *ResNet18* Partial Fine-tuning with Augmentation | 70.96% |
| *ResNet18* Full Fine-tuning with Augmentation | 73.90% |

**Table 2. Comparison of *ResNet18* Test Performance in Different Setups.**

When looking deeper into the prediction for the fully fine-tuned trained model with data augmentation, we realized the model was only predicting one of the classes correctly, namely "normal" stool (see Table 3). This is consistent with the fact that most of the training set is composed of "normal" class instances and therefore the algorithm is biased against this particular class.

| | **Predicted** | constipated | normal | loose |
| --- | --- | --- | --- | --- |
| **Ground Truth** | constipated | 0 | 12 | 0 |
| | normal | 0 | 201 | 0 |
| | loose | 0 | 59 | 0 |

**Table 3. Confusion Matrix for *ResNet18* Fully Fine-tuned with Data Augmentation on the Test Set of 272 images.**

As a comparison, we ran the same model setups through our secondary dataset (i.e. the synthetic stool images made with play doh). Even though the model is clearly overfitting due to low entropy in the secondary dataset, we consider this proof that with a sufficiently large dataset, we will be able to increase the accuracy of the classification task in real samples (see Table 4).

| Model Setup | Accuracy |
| --- | --- |
| *ResNet18* Partial Fine-tuning | 98.8% |
| *ResNet18* Full Fine-tuning | 1.00% |
| *ResNet18* Partial Fine-tuning with Augmentation | 94.4% |
| *ResNet18* Full Fine-tuning with Augmentation | 99.4% |

**Table 4. Comparison of *ResNet18* Test Performance in Different Setups and Trained on the Secondary Dataset.**

## ACKNOWLEDGMENTS OF MENTORSHIP
We thank Professor Deborah Estrin for her continuous support and invaluable advice during this project. We are grateful to Dr. Kyle Staller and Dr. Christopher Velez for their instrumental input regarding the medical aspects of this project as well as the time spent annotating data. We also thank all the anonymous volunteers who provided training data. Finally, we thank Neta Tamir for her support and guidance.